\documentclass[preprint,12pt]{elsarticle}

\usepackage{amssymb}
\usepackage{multirow}
\usepackage{tabularx}
\usepackage{amssymb}
\usepackage{wrapfig} 
\usepackage{graphicx}
\usepackage{amsmath}
\usepackage{xurl}
\usepackage[colorlinks=true, allcolors=blue]{hyperref}
\usepackage{caption}
 
\begin{document}
\begin{sloppypar}
\begin{frontmatter}



\title{Medical Scene Reconstruction and Segmentation based on 3D Gaussian Representation}
\author[label1]{Bin Liu}
\author[label1]{Wenyan Tian}
\author[label1]{Huangxin Fu}
\author[label2]{Zizheng Li}
\author[label1]{Zhi-Fen He}
\author[label1]{Bo Li\corref{c1}}
\affiliation[label1]{
  organization={Nanchang Hangkong University},
  addressline={696 Fenghe South Avenue, Honggutan District},
  city={Nanchang},
  postcode={330063},
  state={Jiangxi},
  country={China}}

\affiliation[label2]{organization={Guilin University Of Electronic Technology},
  addressline={1 Jinji Road, Qixing District},
  city={Guilin},
  postcode={541004},
  state={Guangxi},
  country={China}
}

\cortext[c1]{Corresponding author.}

\begin{abstract}
3D reconstruction of medical images is a key technology in medical image analysis and clinical diagnosis, providing structural visualization support for disease assessment and surgical planning. Traditional methods are computationally expensive and prone to structural discontinuities and loss of detail in sparse slices, making it difficult to meet clinical accuracy requirements.To address these challenges, we propose an efficient 3D reconstruction method based on 3D Gaussian and tri-plane representations. This method not only maintains the advantages of Gaussian representation in efficient rendering and geometric representation but also significantly enhances structural continuity and semantic consistency under sparse slicing conditions. Experimental results on multimodal medical datasets such as US and MRI show that our proposed method can generate high-quality, anatomically coherent, and semantically stable medical images under sparse data conditions, while significantly improving reconstruction efficiency. This provides an efficient and reliable new approach for 3D visualization and clinical analysis of medical images.
\end{abstract}




\end{frontmatter}


\section{Introduction}
\label{sec:introduction}
 Ultrasound, MRI, CT and other medical imaging technologies are the core means of non-invasively obtaining information about human anatomical structures. Medical image three-dimensional reconstruction and semantic segmentation technology can integrate discrete two-dimensional slices into spatially consistent and anatomically coherent three-dimensional models, realize the precise localization of tissues and lesions, and provide important support for clinical diagnosis, preoperative planning and treatment evaluation. It is one of the important research directions in the field of medical artificial intelligence\cite{1}.

Traditional medical image 3D reconstruction methods mostly rely on interpolation strategies or voxel stacking methods to generate 3D volume data by filling grayscale information between adjacent slices. Such methods are simple to implement, but when the slice spacing is uneven or the sampling is sparse, it is easy to introduce problems such as inter-slice discontinuity and blurred tissue boundaries, which leads to limited geometric accuracy and anatomical consistency of the reconstructed structure. At the same time, voxel representation will bring significant storage and computational overhead in high-resolution medical imaging scenarios\cite{2}. In addition, the traditional reconstruction process is usually independent of the semantic segmentation task, lacking a unified 3D semantic modeling capability, making it difficult to support the accurate separation and independent reconstruction of different anatomical structures.

With the development of deep learning technology, data-driven methods have shown significant advantages in the three-dimensional reconstruction of medical images. Deep learning can learn spatial priors and anatomical structure rules from large-scale medical image data, and can more effectively recover complex three-dimensional structural details while reducing noise sensitivity. In recent years, 3DGS\cite{3} has emerged as an explicit and differentiable three-dimensional representation, which has both efficient optimization and high-precision geometric modeling capabilities . It has been introduced into medical image reconstruction and developed into a variety of improved frameworks\cite{4} to alleviate problems such as data sparsity and motion artifacts. However, most existing 3DGS-based medical reconstruction methods focus on projection imaging scenarios such as CT, and the modeling process is highly dependent on external perspective information, making it difficult to directly adapt to in vivo imaging modes such as MRI and fMRI that lack perspective information. At the same time, Gaussian parameters are usually optimized as independent variables, lacking unified global continuous constraints, which leads to structural instability and semantic inconsistency under sparse slice conditions. This limits its application scope in the field of multimodal medical imaging.

To address the aforementioned issues, this paper proposes a hybrid explicit-implicit representation method based on 3D Gaussian and 3D planes for 3D reconstruction of medical images.This method uses tri-plane as explicit, continuous global feature fields to encode spatial structure and semantic context, and maps them to the properties of 3D Gaussian through an implicit decoder, thereby establishing a unified, differentiable, and spatially consistent 3D representation among discrete Gaussian units. This hybrid explicit-implicit modeling approach not only retains the advantages of 3DGS in efficient rendering and explicit geometric representation, but also effectively compensates for the lack of global constraints in discrete Gaussian by utilizing the continuous feature fields of the tri-plane. This makes the modeling from 2D slices to continuous 3D anatomical structures more stable and reliable, achieving structurally coherent, detailed, and semantically consistent 3D medical reconstruction results even under sparse slice conditions.

In summary, the main contributions of our method are as follows :
\begin{enumerate}
    \item A medical 3D reconstruction and segmentation framework based on a hybrid explicit-implicit representation of 3D Gaussian and tri-plane is proposed, which realizes efficient modeling from sparse 2D slices to continuous 3D anatomical structures.
    \item An improved 3D Gaussian model was constructed, which can simultaneously capture the texture and semantic information of anatomical structures, providing a reliable representation for semantically perceptual 3D reconstruction.
    \item A tri-plane global continuity constraint is introduced to enhance the spatial consistency and semantic stability of Gaussian properties, enabling high-quality, anatomically coherent and semantically consistent 3D reconstruction under sparse slicing conditions.
\end{enumerate}

\section{Related Work}
\label{sec:related work}

\subsection{Three-dimensional medical image reconstruction}
3D medical image reconstruction is to recover the 3D spatial structure of internal human tissues from 2D tomographic slices. Based on different representation methods, it can be divided into three categories: voxel representation, mesh representation, and implicit representation.

Voxel representation discretizes three-dimensional space into a regular voxel grid and reconstructs it by assigning density and color values to each voxel. This method is intuitive and easy to understand, yet there is a contradiction between resolution and computational cost. High-resolution voxel grids will bring huge memory overhead.

Mesh representation constructs a 3D model by piecing together triangular facets, which can generate surface models with clear topological structures and is widely used in organ morphology analysis. But this method is difficult to handle complex internal textures and has poor adaptability to sparse slice data.

Implicit representation methods, represented by NeRF\cite{5}, implicitly encode the density and color information of three-dimensional space through neural networks, and can generate continuous and high-precision reconstruction results. In recent years, it has been widely used in the medical field, such as MedNeRF\cite{6}, Ultra-NeRF\cite{7} and RapidVol\cite{8}. However, these methods typically rely on dense sampling strategies for volume rendering, requiring high levels of supervision density or multi-view information, resulting in significant training and inference costs. Furthermore, they struggle to balance reconstruction efficiency and stability under sparse slice conditions, limiting their application in efficient clinical analysis. This paper introduces explicit three-dimensional Gaussian representation to replace pure implicit volume field modeling, which significantly reduces the dependence on slice density and volume rendering sampling while maintaining continuity.

\subsection{Three-dimensional Gaussian splattering and its medical applications}
 
3DGS is an explicit 3D representation method based on differentiable Gaussian units, achieving fast rendering through efficient rasterization. Compared to NeRF , 3DGS significantly improves training and inference speed while maintaining reconstruction accuracy, and has become an important research direction in the field of neural rendering in recent years.

In the field of medical image reconstruction, researchers have made targeted improvements to the characteristics of 3DGS, covering multiple imaging modalities such as ultrasound and MRI. UltraGauss\cite{9} proposed a Gaussian sputtering framework for ultrasound imaging, which extends the new perspective synthesis technology to the ultrasound field by explicitly modeling the probe plane. 3DGR-CT\cite{10} uses a learnable three-dimensional Gaussian set as a voxel representation, and achieves high-fidelity density field recovery under sparse perspective by optimizing Gaussian parameters to fit sparse projection. MedGS\cite{11} represents medical images as the distribution of continuous two-dimensional frames in three-dimensional space, and combines Gaussian modeling and interpolation mechanisms to achieve semi- supervised implicit surface reconstruction. InnerGS\cite{12} is designed for internal scenes in medical images. It uses three-dimensional Gaussian differentiable volume modeling and optimizes the rasterization process to generate slice images along any axis, thus meeting the needs of medical image reconstruction without viewpoint. However, the Gaussian properties in this method are mainly obtained through independent parameter optimization, lacking a unified global continuous constraint, and do not explicitly introduce anatomical semantic information. Under sparse supervision, structural instability and semantic expression are still prone to occur. To address the aforementioned issues, this paper introduces a tri-plane global feature field combined with implicit decoding to modulate Gaussian properties, thereby achieving consistent modeling of structure and semantics while maintaining the efficiency of 3DGS.

\section{Main Method}
\label{sec:main method}
In this section, we will briefly introduce the proposed method for medical image reconstruction and segmentation that combines Gaussian representation with tri-plane.

\textbf{3D Gaussian representation}  3DGS represents the entire scene as a collection of 3D Gaussian primitives \(\left\{\left.G_i\right\}\right._{i=1}^N\), where each Gaussian primitive is defined by the following parameters: \(G_i=\left\{\mu_i\right.,\mathrm{\Sigma}_i,\alpha_i,c_i,\left.s_i\right\}\). Specifically, \( \mu_i\in R^3\) is the Gaussian mean, \ \(\mathrm{\Sigma}_i\in R^{3\times3}\) is the anisotropic covariance matrix used to characterize the scale and shape of local structures in different directions , \ \(\alpha_i\) is the opacity used to model the contribution of voxels or tissues to light occlusion , \(c_i\) and is the spherical harmonic coefficient used to simulate view-dependent color or intensity changes. The influence of a single Gaussian primitive \( G_i(x)=e^{-\frac{1}{2}{(x-\mu_i)}^T\Sigma_i^{-1}(x-\mu_i)}\) at any point in space is represented as: \(x\in R^3\).

Traditional 3DGS is based on projection models, but medical images such as MRI lack an external perspective, making projection rendering unsuitable. This study introduces the InnerGS internal scene rasterizer to transform rendering into in vivo slice imaging, utilizing the interpretability and local expressive power of 3D Gaussians to adapt to the needs of fine structure reconstruction.

To achieve joint modeling of geometry and semantics, we add a semantic feature parameter \(s_i\) to each Gaussian unit, so that it simultaneously possesses geometric, color and semantic information. When rendering the real image and semantic feature image on the slice plane at a specified axis and depth\( t \), we no longer rely on projection methods, but instead obtain the final result by accumulating the contribution of each 3D Gaussian \( G_i\) to the pixel \(G_i\left(u,\ v,\ t\right)\), thus achieving joint supervision of the geometric, color, and semantic aspects of the Gaussian distribution. The calculation formulas for real slice pixels \(I\left(u,v,t\right)\) and semantic slice pixels \(S\left(u,v,t\right)\) are as follows :

\begin{equation}
I\left(u,v,t\right)=\sum_{i=1}^{N}p_i(u,v,t)\alpha_ic_i\prod_{j=1}^{i-1} (1-p_j(u,v,t)\alpha_j)    
\end{equation}
\begin{equation}
S\left(u,v,t\right)=\sum_{i=1}^{N}p_i(u,v,t)\alpha_is_i\prod_{j=1}^{i-1} (1-p_j(u,v,t)\alpha_j)
\end{equation}

To improve computational efficiency, InnerGS decomposes the 3D Gaussian along the slice normal direction into the product of a 1D Gaussian and a 2D conditional Gaussian: where \ \(G_i\left(u,v,t\right)=G_i\left(u,v|t\right)G_i\left(t\right)\)) describes \(G_i\left(u,v|t\right)\) the lateral distribution within the slice plane, and describes the uncertainty along the depth axis. This strategy can efficiently calculate the effective range of the Gaussian on a single slice, significantly reducing \( G_i\left(t\right)\) computational complexity and supporting fast rendering and end-to-end optimization of high-resolution slices.

\textbf{Tri-plane }  When modeling using only unordered Gaussian sets, the lack of global contextual relationships among Gaussian elements can easily lead to incoherent cross-slice structures and insufficient semantic consistency. To address this issue while considering storage and computational efficiency, this paper decomposes the 3D spatial features into three orthogonal 2D planes aligned along the axes , resulting in \(F_{xy}\left(x,y\right),F_{yz}\left(y,z\right),F_{xz}\left(x,z\right)\). For a spatial point \( p=(x,y,z)\), it is projected onto tri-plane, sampled, and its features are fused to obtain:\(f\left(p\right)=\varphi\left(F_{xy}\left(x,y\right),F_{yz}\left(y,z\right),F_{xz}\left(x,z\right)\right)\) , where \(\varphi\)  represents the feature concatenation or weighted summation operation.

The tri-plane feature field, as an explicit global feature store, provides a continuous implicit representation for each point in space by sampling on tri-plane and decoding with an MLP, thereby creating a consistency constraint on the cross-slice structure. In 3D reconstruction of medical images, thetri-plane naturally align slices along each axis (axial, coronal, sagittal), allowing each Gaussian unit to obtain continuous contextual information on slices in different directions, improving the coherence of the cross-slice structure and the accuracy of tissue distribution. This design also significantly reduces computational complexity and memory overhead: voxel grid storage is reduced from \(O(n^3)\) to \(O(n^2)\), and feature queries only require two-dimensional bilinear interpolation plus lightweight MLP decoding, without traversing the complete 3D voxels, thus accelerating training and inference speeds.

\begin{figure}
    \centering
    \includegraphics[width=0.95\linewidth]{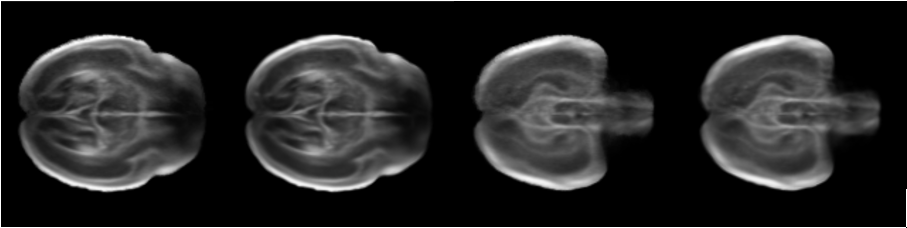}
    \caption{The results of two groups of fetal brain reconstructions from new perspectives. The left image is the ground truth, and the right image is the reconstructed image.}
    \label{fig:placeholder}
\end{figure}

\begin{table}
    \centering
    \begin{tabular}{|c|l|c|c|l|}\hline
         Input Set&method&  SSIM($\uparrow$)&  PSNR($\uparrow$)&LPIPS($\downarrow$) \\\hline
         25\% Axial&InnerGS &  0.9664 ± 0.0236
&  34.74 ± 9.94& 0.0740 ± 0.0481\\
  Slices& Ours& 0.9812 ± 0.0139& 37.61 ± 6.23&0.0433 ± 0.0194\\\hline
         50\% Axial &InnerGS 
& 0.9851 ± 0.0099& 38.14 ± 9.37& 0.0434 ± 0.0283\\
 Slices& Ours& 0.9947 ± 0.0034& 41.44 ± 2.83&0.0255 ± 0.0131\\\hline
 95\% Axial &InnerGS 
& 0.9876 ± 0.0085& 38.00 ± 6.21& 0.0371 ± 0.0201\\
 Slices& Ours& 0.9963  ±  0.0026& 41.90 ± 1.50&0.0208 ± 0.0126\\\hline
    \end{tabular}
    \caption{In the sparse slice experiment, the average SSIM, PSNR, and L PIPS were reconstructed on the test set  }
    \label{tab:placeholder}
\end{table}

\section{Experiments}
\label{sec:experiments}
\subsection{Sparse slice reconstruction }

We used fetal brain ultrasound volume data from INTERGROWTH-21st\cite{16} for sparse slice reconstruction experiments. The volume data was 160×160×160 voxels.160 slices were sampled along the volume data axis, with each slice having a resolution of 160×160 voxels. In the experiment, we selected 25\%, 50\% and 95\% of the slices as the training set, and the remaining slices as the test set.

Figure 1 shows the results of the novel perspective reconstruction of the fetal brain under the 50\% slice training condition, where the left image is the ground truth and the right image is the reconstructed image using the proposed method. It can be seen that under sparse slice conditions, the novel perspective image generated by this method is highly consistent with the real slices, and the complex structure of the fetal brain can be accurately reconstructed, demonstrating the ability of this method to maintain structural coherence and detail fidelity under sparse data.

Table 1 shows the experimental results, demonstrating that the proposed method consistently outperforms InnerGS under different axial slice sampling ratios. The performance difference is most pronounced under the extremely sparse 25\% slice condition: compared to the baseline method, this method improves SSIM from 0.9664 to 0.9812, PSNR from 34.74 dB to 37.61 dB, and LPIPS from 0.0740 to 0.0433, proving that this method can effectively maintain 3D structural continuity and perceived quality even with very few observations. As the training slice ratio increases to 50\% and 95\%, the performance gap between the two methods narrows, but this method still maintains a stable advantage across all metrics, demonstrating better detail fidelity and cross-slice consistency.

\begin{table}
    \centering
    \begin{tabular}{|c|c|c|l|}\hline
         axis&  PSNR& SSIM&Semantic loss\\\hline
         Axial&  35.78 ± 3.16&  0.9616 ± 0.0140&0.0133 ± 0.0051\\\hline
         Coronal&  29.45 ± 3.86&  0.8627 ± 0.0851&0.0022 ± 0.0025\\\hline
 Sagittal& 32.04 ± 5.02& 0.9437 ± 0.0643&0.0046 ± 0.0063\\ \hline
    \end{tabular}
    \caption{ The average PSNR, SSIM, and semantic image average loss metrics for the prostate dataset across the three axes   }
    \label{tab:placeholder}
\end{table}

Overall, this method not only has significant advantages in sparse slice scenarios, but also continuously improves reconstruction quality as slice density gradually increases, demonstrating stronger robustness and the ability to effectively characterize the internal continuity of ultrasound body data.

\begin{figure}
    \centering
    \includegraphics[width=0.95\linewidth]{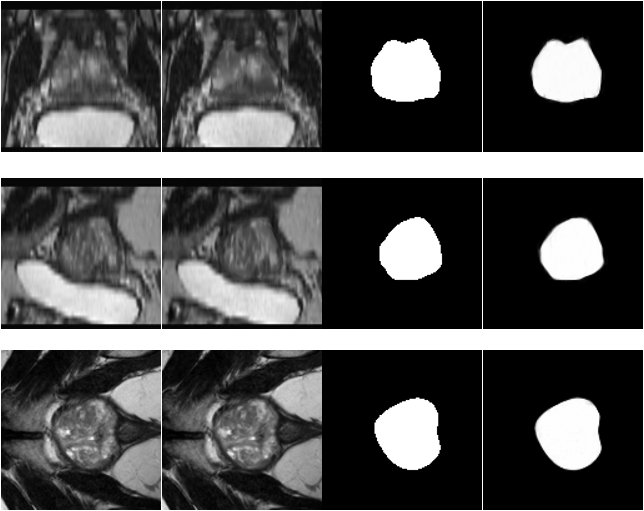}
    \caption{The reconstruction results of the prostate dataset on three axes, including images and semantics. The left image is the ground truth, and the right image is the reconstructed image.
}
    \label{fig:placeholder}
\end{figure}

\begin{table}
    \centering
    \begin{tabular}{|c|c|c|l|}\hline
         axis&  PSNR& SSIM&Semantic loss
\\\hline
         A xial&  35.78 ± 3.16& 0.9616 ± 0.0140&0.0133 ± 0.0051\\\hline
         Coronal
&  41.05 ± 1.97&  0.9798 ± 0.0037&0.0118 ± 0.0095\\\hline
 S agittal& 35.07 ± 5.04& 0.9656 ± 0.0114&0.0137 ± 0.0088\\ \hline
    \end{tabular}
    \caption{Average PSNR, SSIM, and semantic image average loss of reconstructed images. }
    \label{tab:placeholder}
\end{table}
\subsection{Semantic segmentation}

We conducted a new perspective semantic segmentation experiment using two datasets.The prostate MRI dataset\cite{17} is the publicly available prostate MRI dataset from the MICCAI 2023 MRI to Ultrasound Prostate Registry Challenge, which contains prostate data from 108 patients. Each case includes clearly annotated anatomical landmarks.

Table 2 shows that the proposed method exhibits good reconstruction performance and semantic segmentation accuracy across all three anatomical axes of the prostate MRI dataset. The axial axis performs best, with a PSNR of 35.78 dB and an SSIM of 0.9616, demonstrating accurate restoration of axial anatomical details. The semantic loss is controlled at 0.0133 ± 0.0051 , indicating stable and reliable axial semantic segmentation results. The sagittal axis is second best; although there are some fluctuations in the indicators, overall high structural consistency and semantic accuracy are maintained. The coronal axis has the lowest PSNR and SSIM values among the three axes, presumably related to the higher complexity of details and increased blurring of tissue boundaries in the coronal anatomical structures during imaging. However, the semantic loss is significantly lower than the other two axes, indicating a significant advantage in semantic segmentation accuracy in this direction, and the model can effectively capture the anatomical semantic features of the coronal plane.

Combining the visualization results in Figure 2, the images and semantic segmentation maps reconstructed by the proposed method are highly consistent with the real data, clearly restoring the anatomical boundaries of the prostate and accurately dividing the semantic regions, thus verifying the effectiveness of the method in the new perspective reconstruction and semantic segmentation tasks of prostate MRI.

The HVSMR-2.0 dataset\cite{18} provides three-dimensional cardiovascular magnetic resonance images and corresponding anatomical mask images. Each case records the static anatomical structure of the entire heart. We extract two-dimensional slices and two-dimensional grayscale semantic maps along three anatomical directions and map the grayscale semantic maps to RGB semantic maps.

\begin{figure}
    \centering
    \includegraphics[width=0.95\linewidth]{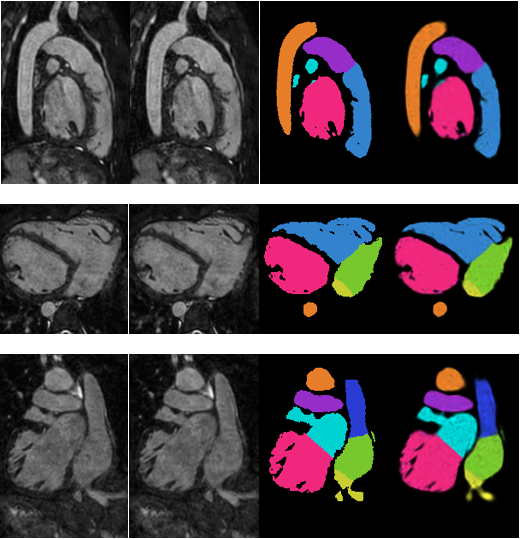}
    \caption{The reconstruction results of cardiac MRI images and semantics on three axes. The left image is the ground truth, and the right image is the reconstructed image.
}
    \label{fig:placeholder}
\end{figure}

The experimental results in Table 3 and Figure 3 demonstrate that the proposed method also exhibits excellent performance in cardiac MRI imaging scenarios. The reconstructed images not only completely restore the overall morphology and internal anatomical structures of the heart, but the semantic segmentation map can also accurately distinguish different tissues and anatomical regions, showing a high degree of consistency with the actual annotations. This indicates that the proposed method can maintain structural continuity and semantic consistency under sparse slicing conditions, while supporting semantically perceptual reconstruction from a new perspective, fully validating its modeling ability and adaptability to complex cardiovascular anatomy.

\section{Conclusion}
\label{sec:conclusion}
This paper addresses the limitations of traditional 3DGS in reconstructing internal anatomical structures in medical images, including limited adaptability, insufficient tissue boundary delineation, and lack of semantic segmentation capabilities. It proposes a framework for 3D reconstruction of medical images that integrates a three-plane feature field and a 3D Gaussian representation. This method introduces semantic feature attributes into the 3D Gaussian representation and utilizes explicit global features from the three planes for implicit decoding, achieving joint modeling of geometric, appearance, and semantic information. This effectively improves the continuity and semantic consistency of structures across slices. Experimental results based on fetal brain ultrasound, prostate MRI, and cardiac MRI datasets demonstrate that this method outperforms existing methods in terms of PSNR, SSIM, and LPIPS, achieving high-quality internal anatomical structure reconstruction and accurate semantic segmentation even under sparse slice conditions. This research provides an efficient and reliable solution for 3D visualization and clinical diagnosis of medical images, possessing significant clinical application value and potential for widespread adoption.

\section*{Acknowledgements}
The work is partially funded by Natural Science Foundation of China (62172198, 61762064, 62362051), Academic Leaders Training Program of Jiangxi Province  (20232BCJ22001), Key Project of Jiangxi Natural Science Foundation (20224ACB202008), Key R\&D Plan of Jiangxi Province (20232BBE50022), Jiangxi Provincial Natural Science Foundation (20232BAB212013, 20232BAB202047) and the Opening Project of Nanchang Innovation Institute, Peking University.
\bibliographystyle{elsarticle-num}


\bibliography{reference}

\end{sloppypar}

\end{document}